\documentclass[letterpaper, 10 pt, conference]{ieeeconf}  % Comment this line out if you need a4paper

\IEEEoverridecommandlockouts                              % This command is only needed if 
\usepackage{graphicx}
\usepackage{amsmath} % assumes amsmath package installed
\usepackage{amssymb}  % assumes amsmath package installed

\usepackage{subcaption}
\usepackage{array}
\usepackage{flushend}
\usepackage{cite}   % collect references numbers inside single set of square brackets

\usepackage{amsmath, amssymb}

\usepackage{xifthen}
\usepackage{color}

\newcommand{\xstrut}[1]{\rule{0ex}{#1ex}}

\newcommand{\ba}{\begin{eqnarray}}
\newcommand{\ea}{\end{eqnarray}}
\newcommand{\eq}[1]{(\ref{#1})}

\newcommand{\unit}[1]{\mbox{$\mathrm{\,#1}$}}

\newcommand{\R}{\mathbb{R}}

\newcommand{\presup}[1]{\,{}^{\scriptscriptstyle #1}\!}

\newcommand{\pose}[1][ZZZZ]{\ifthenelse{\equal{#1}{ZZZZ}}{}{\presup{#1}}{\mathbf{\xi}}}
\newcommand{\estpose}[1][ZZZZ]{\ifthenelse{\equal{#1}{ZZZZ}}{}{\presup{#1}}{\mathbf{\hat{\xi}}}}
\newcommand{\hpose}[1][ZZZZ]{\ifthenelse{\equal{#1}{ZZZZ}}{}{\presup{#1}}{\hat{\mathbf{\xi}}}}
\newcommand{\posedot}[1][ZZZZ]{\ifthenelse{\equal{#1}{ZZZZ}}{}{\presup{#1}}{\mathbf{\nu}}}
\newcommand{\q}[1][ZZZZ]{\ifthenelse{\equal{#1}{ZZZZ}}{}{\presup{#1}}{\mathring{q}}}
\DeclareMathAlphabet{\mathitbf}{OML}{cmm}{b}{it}
\newcommand{\twist}[2][ZZZZ]{\ifthenelse{\equal{#1}{ZZZZ}}{}{\presup{#1}}{\mathcal{S}}}
\renewcommand{\vec}[2][ZZZZ]{\ifthenelse{\equal{#1}{ZZZZ}}{}{\presup{#1}}{\mathitbf{#2}}}
\newcommand{\hvec}[2][ZZZZ]{\ifthenelse{\equal{#1}{ZZZZ}}{}{\presup{#1}}{\tilde{\vec{#2}}}}
\newcommand{\evec}[2][ZZZZ]{\ifthenelse{\equal{#1}{ZZZZ}}{}{\presup{#1}}{\hat{\vec{#2}}}}
\newcommand{\bvec}[2][ZZZZ]{\ifthenelse{\equal{#1}{ZZZZ}}{}{\presup{#1}}{\bar{\vec{#2}}}}
\newcommand{\dvec}[2][ZZZZ]{\ifthenelse{\equal{#1}{ZZZZ}}{}{\presup{#1}}{\dot{\vec{#2}}}}
\newcommand{\ddvec}[2][ZZZZ]{\ifthenelse{\equal{#1}{ZZZZ}}{}{\presup{#1}}{\ddot{\vec{#2}}}}

\newcommand{\mat}[2][ZZZZ]{\ifthenelse{\equal{#1}{ZZZZ}}{}{\presup{#1}\,}{{\boldsymbol #2}}}
\newcommand{\dmat}[2][ZZZZ]{\ifthenelse{\equal{#1}{ZZZZ}}{}{\presup{#1}\,}{{\dot{\boldsymbol #2}}}}
\newcommand{\emat}[2][ZZZZ]{\ifthenelse{\equal{#1}{ZZZZ}}{}{\presup{#1}\,}{\hat{\boldsymbol#2}}}
\newcommand{\matfn}[3][ZZZZ]{\ifthenelse{\equal{#1}{ZZZZ}}{}{\presup{#1}}{{\mat{#2}}\left(#3\right)}}
\newcommand{\Rt}[2][ZZZZ]{\ifthenelse{\equal{#1}{ZZZZ}}{}{\presup{#1}}{{\bf R}\left(#2\right)}}
\newcommand{\cframe}[1]{\{#1\}}

\newcommand{\point}[2][ZZZZ]{\ifthenelse{\equal{#1}{ZZZZ}}{}{\presup{#1}}{\mathbf{\mathrm{#2}}}}

\newfont{\School}{pncr}
\newfont{\eightTR}{pncr at 8pt}
\usepackage{color}
\usepackage{fancyvrb}
\fvset{formatcom=\color{blue},fontseries=c,fontfamily=courier,xleftmargin=4mm,commentchar=!}
\DefineVerbatimEnvironment{Code}{Verbatim}{formatcom=\color{blue},fontseries=c,fontfamily=courier,fontsize=\footnotesize,xleftmargin=4mm,commentchar=!}
\DefineVerbatimEnvironment{CodeSmall}{Verbatim}{formatcom=\color{blue},fontseries=c,fontfamily=courier,fontsize=\scriptsize,xleftmargin=1mm,commentchar=!}
\DefineVerbatimEnvironment{CodeNum}{Verbatim}{numbers=left,numbersep=4pt,formatcom=\color{blue},fontseries=c,fontfamily=courier,fontsize=\footnotesize,xleftmargin=4mm}
\newcommand{\model}[1]{\index{code}{#1@\textit{#1}}\ifthenelse{\boolean{draft}}{{\color{green}\Verb+#1+}}{\Verb+#1+}}
\newcommand{\block}[1]{\ifthenelse{\boolean{draft}}{{\color{green}\Verb+#1+}}{\textsf{#1}}}
\newcommand{\func}[2][ZZZZ]{\ifthenelse{\equal{#1}{ZZZZ}}{\index{code}{#2}}{\index{code}{#1}}\ifthenelse{\boolean{draft}}{{\color{green}\Verb+#2+}}{\Verb+#2+}}
\newcommand{\methodb}[2]{\index{code}{#1@\textbf{#1}!.#2}\ifthenelse{\boolean{draft}}{{\color{magenta}\Verb+#1.#2+}}{\Verb+#1.#2+}}
\newcommand{\method}[2]{\index{code}{#1@\textbf{#1}!.#2}\ifthenelse{\boolean{draft}}{{\color{magenta}\Verb+#2+}}{\Verb+#2+}}
\newcommand{\class}[1]{\index{code}{#1@\textbf{#1}}\ifthenelse{\boolean{draft}}{{\color{cyan}\Verb+#1+}}{\Verb+#1+}}
\newcommand{\property}[1]{\index{property}{#1}\ifthenelse{\boolean{draft}}{{\color{cyan}\Verb+#1+}}{\Verb+#1+}}
\makeatletter
\newcommand*{\balancecolsandclearpage}{%
  \close@column@grid
  \clearpage
  \twocolumngrid
}
\makeatother

\title{\LARGE \bf
Control of the Final-Phase of Closed-Loop Visual Grasping using Image-Based Visual Servoing
}

\author{Jesse Haviland$^{1}$, Feras Dayoub$^{1}$, Peter Corke$^{1}$% <-this % stops a space
\thanks{$^{1}$Jesse Haviland, Feras Dayoub, and Peter Corke are with the Australian Centre for Robotic Vision (ACRV), Queensland University of Technology (QUT), Brisbane, Australia
        {\tt\small j.haviland@qut.edu.au, feras.dayoub@qut.edu.au, peter.corke@qut.edu. au}. This research was conducted by the Australian Research Council project number CE140100016, and supported by the QUT Centre for Robotics.
}%
}

\begin{document}

\maketitle
\thispagestyle{empty}
\pagestyle{empty}

%%%%%%%%%%%%%%%%%%%%%%%%%%%%%%%%%%%%%%%%%%%%%%%%%%%%%%%%%%%%%%%%%%%%%%%%%%%%%%%%
\begin{abstract}
This paper considers the final approach phase of visual-closed-loop grasping where the RGB-D camera is no longer able to provide valid depth information. 
Many current robotic grasping controllers are not closed-loop and therefore fail for moving objects.
Closed-loop grasp controllers based on RGB-D imagery can track a moving object, but fail when the sensor's minimum object distance is violated just before grasping.
To overcome this we propose the use of image-based visual servoing (IBVS) to guide the robot to the object-relative grasp pose using camera RGB information.  IBVS robustly moves the camera to a goal pose defined implicitly in terms of
an image-plane feature configuration. In this work, the goal image feature coordinates are \textit{predicted} from RGB-D data to enable  RGB-only tracking once depth data becomes unavailable -- this enables more reliable grasping of previously unseen moving objects.
Experimental results are provided.

\end{abstract}

%%%%%%%%%%%%%%%%%%%%%%%%%%%%%%%%%%%%%%%%%%%%%%%%%%%%%%%%%%%%%%%%%%%%%%%%%%%%%%%%

\section{Introduction}

Autonomous and reliable grasping is crucial to robots performing useful tasks in the real world. 
A robust robotic grasper must have the ability to: operate in situations where objects may be moving; be robust to errors in sensing or actuation; grasp items that  have never been seen previously.

\begin{figure}[t]
    \centering
    \includegraphics[ height=4.9cm]{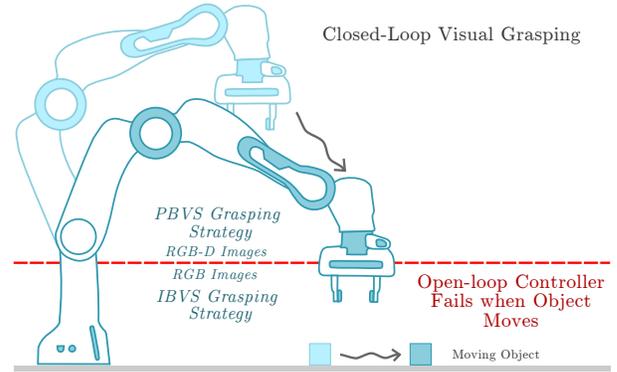}
    \caption{
        During the final approach phase  depth information is no longer available from the RGB-D sensor. We use image-based visual servoing for the final motion phase and servo toward a goal feature configuration predicted from the last valid depth image.
    }
    \vspace{-14pt}
    \label{fig:cover}
\end{figure}

To grasp robustly, with respect to dynamic scenes or sensor/actuation error, a closed-loop approach is required with the ability to perform grasp synthesis at a sufficient rate to use in the control loop. For example, the system \cite{doug}, which is extended in this work, provides a grasp pose in just 19\unit{ms} given a depth image of the scene. However, all RGB-D cameras have a minimum sensing distance for range data, typically in the order of 30\unit{cm} as illustrated in Figure \ref{fig:cover}. For objects closer than this ``standoff'' distance the camera provides an RGB image but no valid depth data.  

This means that a dynamic grasp planner such as \cite{doug} will fail during the final grasp phase if the object is still moving. 
However,  RGB cameras can generally operate reliably at close range subject to constraints on depth of field, field of view and occlusion.

In order to perform closed-loop control below the minimum depth sensing distance of the RGB-D camera, a visual servoing (VS) scheme is proposed. Image-based visual servoing (IBVS) is a control approach that uses a set of image-plane visual features (point coordinates \cite{vs1}, parameters of lines or ellipses \cite{vs0}, or image moments\cite{vs4,vs3}) to guide a camera to a desired pose with respect to the scene\cite{vs1,vs2}. A particular advantage of IBVS, compared to other VS schemes \cite{vs1}, for point-coordinate features is that the points are driven in straight lines on the image-plane and never leave the field of view. 

IBVS is simple and remarkably robust but in practice there are three challenges:
we need to know the distance from the camera to the object; we need to establish robust correspondence between the current and the goal features; and we need to know the goal feature configuration.
Firstly, in our case object distance can be inferred from a valid depth image and subsequent robot joint-encoder odometry.
Secondly, many techniques exist for robustly establishing correspondence and we propose the use of scale- and rotation-invariant features.
Thirdly, goal feature configuration can be \textit{predicted} from a valid depth image, combined with current and desired end-effector pose and is a key contribution of this work.

Further details on each of these system components are provided in the remainder of this paper.
We assume that the object motion is planar, that the RGB camera can provide a focussed image at close range, and that
the robot's fingers do not occlude the object.

The contributions of this paper are:
\begin{enumerate}
    \item The use of IBVS to extend the closed-loop working range of depth-image-based grasping controller so as to allow more robust grasping of moving objects.
    \item A novel method to predict the goal image-feature configuration for IBVS, from an RGB-D image.
    \item Experimental validation with unmodeled moving objects.
\end{enumerate}
Section \ref{sec:related-work} describes related work, Section \ref{sec:control} describes our control approach, Section \ref{sec:exp} describes our experimental setup and methodology and, finally, Section \ref{sec:results} details our experimental results and insights informed by the results.

\section{Related Work}\label{sec:related-work}

\subsection{Visual Servoing for Grasping}

A key problem in robotic grasping is determining the grasp pose.  Grasp pose synthesis is a well established field with many methods \cite{graspOld1,graspOld3}.  Recently, grasp point synthesizers using deep learning approaches have proven very successful, even on never-before-seen items and scenes \cite{doug,grasp2,grasp3,grasp4}.  These synthesizers have been able to learn effective and important features present in depth images to output reliable grasp poses. 

Closed-loop grasping involves using a VS scheme to position the robot's end-effector in such a way that it can grasp an object. The current state-of-the-art in robotic grasping uses position-based visual servoing (PBVS)  to guide the end-effector to the grasp point \cite{doug, grasp2, grasp3, grasp4} based on the estimated camera-relative pose of the object grasp point.
However, these approaches rely on depth information from an RGB-D camera which has a minimum operating distance, therefore the last stage of the grasp must be completed in an open-loop manner.

\subsection{Minimum sensing distance}

Structured light \cite{sl} and stereo \cite{stereo} cameras exploit multiple view geometry with a fixed transform between two different camera sensors to construct 3D information\cite{hartley}. The transform between the two sensors is known as the baseline. On structured light cameras, the baseline causes the projected pattern to be outside the field-of-view of the camera at the minimum standoff distance. For stereo cameras at close range, the amount of overlap between views is limited and most algorithms enforce a maximum disparity search which is inversely proportional to range.

Time-of-flight (ToF) cameras construct 3D information by measuring the round trip time of a projected light signal \cite{tof}. However, these will not operate at the minimum standoff distance due to the minimum measurable round-trip time for the light pulse emitted by the camera.

\subsection{Using a Feature Detector for Visual Servoing}

Image feature points are a popular choice for IBVS in unstructured environments. Feature descriptors describe a support region around the feature point and are essential for reliable matching across views -- they are ideally invariant to scale, orientation, illumination, and affine transformations.

Common feature detector and descriptor combinations include: SIFT \cite{sift}, SURF \cite{surf}, ORB \cite{orb} and MSER \cite{mser}. 
A study \cite{descriptorStudy} concluded that SIFT descriptors \cite{sift} were the best in all categories other than robustness luminance changes.  Despite the computation time, we have used SIFT in this work.

For IBVS it is critical, at every time step, to locate each goal feature in the current image.
Features  are initially matched between frames based on descriptor distance.
Greater robustness can be achieved by various heuristics such as the ratio test \cite{sift}, loop consistency\cite{lcc}, or epipolar constraints enforced by computing the fundamental matrix \cite{hartley} with Random Consensus Algorithm (RANSAC) \cite{ransac}.

Other approaches exist which variously: use image reference features with SIFT descriptors \cite{siftVS3}, use epipolar lines to define a sliding visual servoing scheme \cite{siftVS4}, apply IBVS on a mobile robot \cite{siftVS5}, and use SIFT features and descriptors with a known 3D model of the goal object to guide a position-based visual servoing (PBVS) scheme \cite{siftVS6}. Deep learning has provided alternatives to feature matching, such as monocular depth estimation, or depth reconstruction \cite{deepf, deepf2}. However, these are computationally expensive and trained on large-scale scenes rather than close-up images. 

The approaches in \cite{siftVS3}, \cite{siftVS4}, and \cite{siftVS5} require prior knowledge of the goal feature configuration. Typically in IBVS this comes from moving the camera to the goal pose but for the problem we are considering this is not possible. 
Instead the goal feature configuration must be estimated, for a previously unseen object, from observed RGB-D data and measured robot pose.

\begin{figure*}[t]
    \centering
    \includegraphics[width=16.2cm, height=4.5cm]{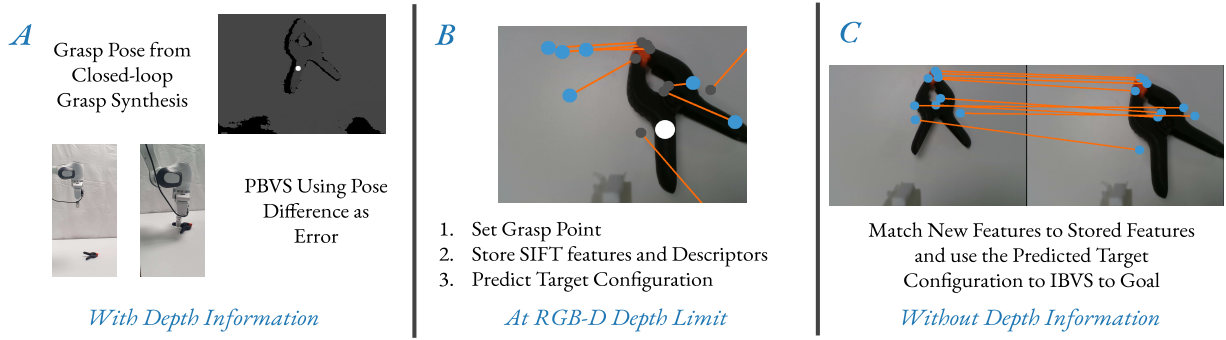} 
    \caption{ Overview of proposed switching grasp controller. The robot continuously performs grasp point synthesis while using a PBVS scheme to approach the grasp point (shown in \textit{A}). At the depth limit of the RGB-D camera the controller sets a grasp pose, stores SIFT key points with locations of the current view of the scene, and predicts the target configuration of these features (shown in \textit{B}). After this point, features in the current scene are matched to the stored information, before using the predicted target configuration of these features to IBVS to the goal (shown in \textit{C}).}
    \label{fig:switching}
\end{figure*}

\section{Proposed Grasp Controller}\label{sec:control}

This section outlines our proposed grasp controller.  The primary sensor is an Intel RealSense D15 RGB-D camera and the control loops run at 30\unit{Hz}. 

The key aspects of our controller are:
\begin{enumerate}
        \item Perform continuous grasp pose synthesis and PBVS to approach the object, utilizing 3D information from an RGB-D camera. See Frame A in Figure \ref{fig:switching} and Section \ref{sec:rgbd-reaching}.
        We also find SIFT image features on the object and record their position, descriptor and depth.
        \item At the lower depth limit of the camera, estimate the image-plane locations of the SIFT features for a camera at the grasp pose. See Frame B in Figure \ref{fig:switching} and Section \ref{sec:goal-estim}.
        \item Below the lower depth limit of the camera, robustly match features in the camera's view to the last stored features and perform IBVS control to drive the former to the latter. See Frame C in Figure \ref{fig:switching} and Section \ref{sec:rgb-reaching}.
        \item Grasp the object.
\end{enumerate}

\textbf{Nomenclature} We use the notation of \cite{peter} where $\cframe{x}$ denotes a coordinate frame, $^x\vec{P} \in \R^3$ is a point in 3D space defined with respect to the coordinate frame $\cframe{x}$, and $\pose[y]_x$ is a relative pose or rigid-body transformation of $\cframe{x}$ with respect to $\cframe{y}$, and $\bullet$ represents composition. Additionally, we use  $^{x}\mat{I} = \R^{H \times W}$ (where $H$ and $W$ are the height and width of the image) to denote an image captured by a camera in the $\cframe{x}$ frame, and $^x\vec{p}  \in \R^5$ to denote an image-plane coordinate of $^x\mat{I}$.  We define coordinate frames `w' for world, `c' for camera, and `e' for end-effector.  The superscript `*' denotes demand.

%%%%%%%%%%%%%%%%%%%%%%%%%%%%%%%%%%%%%%% INITIAL APPROACH
\subsection{RGBD-based initial reaching phase}\label{sec:rgbd-reaching}

\subsubsection{Grasp Pose Calculation}\label{sec:grasp-pose}
We utilize the grasp synthesizer of Morisson et. al. \cite{doug} which, unlike competing deep learning approaches, provides a real-time output (reported as 19 ms per image) making it suitable for a closed-loop system.
It takes a depth image $^{c}\mat{I}_d \in \R^{H \times W}$ and outputs an antipodal grasp encoded as
\[
\vec{G} = ( \vec{\Phi}, \vec{W}, \vec{Q}) \in \R^{3 \times H \times W}
\]
where $\vec{\Phi}, \vec{W}, \vec{Q} \in \R^{H \times W}$
and whose values represent respectively the finger orientation, finger width and grasp quality at every pixel.
The best visible grasp is given by $\vec{g} = \mbox{max}_{\vec{Q}} \vec{G}$ which is a tuple $(s, \phi, w, q)$ where $s = (u,v)$  is the coordinate in $\vec{Q}$ with the greatest grasp quality and $\phi, w, q$ are the corresponding finger orientation, width and grasp quality.
The grasp is expressed in camera image coordinates and the camera axis is
assumed parallel to the table surface.

There is potential for similarly ranked grasps in multiple locations of the image and the grasp point may be unstable which degrades the speed and quality of the control. To counteract this, the previous $\vec{G}^*$ is stored as $\vec{G}_p$ and the next $\vec{G}^*$ is defined as the closest (in image plane coordinates) local maxima around $\vec{G}_p$.

The general form of the camera projection equation for a calibrated camera is
\[ \label{m:camI2}
\begin{pmatrix}
u\\
v\\
w
\end{pmatrix}
 = \mathcal{C}(\pose[0]_c^f)
 \begin{pmatrix}
X\\
Y\\
Z\\
1 \\
\end{pmatrix}
\]
where the left-hand side is the homogeneous image-plane coordinate of the Cartesian point $(X,Y,Z)$
and $\mathcal{C}(\cdot)\in \mathbb{R}^{3 \times 4}$ is the camera matrix, a function of camera pose and intrinsics, 
which we can write in partitioned form
\[
\begin{pmatrix}
\vec{f} \\
w\\
\end{pmatrix}
 = \begin{pmatrix} \mat{A} & \mat{B} & \mat{C} \\
\mat{D} & E & F 
\end{pmatrix}
\begin{pmatrix}
X\\
Y\\
Z\\
1
\end{pmatrix}
\]
where $\mat{A} \in \R^{2\times 2}$, $\mat{B}, \mat{C} \in \R^{2\times 1}$, $\mat{D} \in \R^{1\times 2}$ and the rest scalar.
We can solve for $(X,Y)$ since $\vec{f}$ and $Z$ are known
\begin{equation}
\begin{pmatrix} X\\ Y \end{pmatrix} = 
(\mat{D}^T \vec{f}^T - \mat{A})^{-1} \left[\xstrut{2.3} \mat{B} Z + \mat{C} - (E Z + F) \vec{f} \right] \label{eq:XY}
\end{equation}

The best grasp is
\begin{displaymath}
\vec{G} = (^c\vec{P}_s, ^c\theta_y, W, q)
\end{displaymath}
where $^c\vec{P}_s=(X,Y,Z)$ represents the 3D location of the grasp, $^c\theta_y$ represents the finger orientation (yaw) of the grasp, $W$ represents the grasp width in metres, and $q$ represents the grasp quality.
The grasp pose can be conveniently represented as  Cartesian position and XYZ roll, pitch, and yaw angles 
\begin{displaymath}
\pose[c]_g= 
\begin{pmatrix} 
{^{c}{P}}_{s_x} &
{^{c}{P}}_{s_y} &
{^{c}{P}}_{s_z} &
0 &
0 &
^{c}\theta_y
\end{pmatrix}^T \in \R^6
\end{displaymath}
where roll and pitch angles are $0$ since the fingers are assumed to be normal to the table. 

The fingers can be closed when frame $\cframe{e}$ equals $\cframe{g}$ but pose is measured with respect to the camera which is offset from the end effector by $\pose[e]_c$.
The desired camera pose, with respect to current camera pose, is therefore
\begin{equation} \label{m:posediff}
\pose[c]_{c^*} = {\pose[c]_g} \bullet \pose[e]_c  \,\,\,.
\end{equation}

\subsubsection{PBVS Controller}

The PBVS controller is defined as
\begin{displaymath}
\vec{\nu} =  \vec{\lambda}_p \vec{e}
\end{displaymath}
where $\vec{\nu} = (v_x, v_y, v_z, \omega_x, \omega_y, \omega_z) \in \R^6$ is the end-effector spatial velocity in the world frame, $\vec{\lambda}_p \in \R^{6 \times 6}$ is the diagonal controller gain matrix, and $\vec{e}$ describes the pose error i.e. $\pose[c]_{c^*}$. This controller will run while depth data is available from the RGB-D camera. 

\subsubsection{Finding visual features}
At every time step we extract the $n$-strongest SIFT features from the RGB image, that belong to the objet, and form a list of  reference features and descriptors
\begin{equation}
\vec{\Theta} = \left( (\vec{p}_1, \vec{d}_1, z_1) \hdots, (\vec{p}_n, \vec{d}_n, z_n) \right) \label{eq:features}
\end{equation}
where $\vec{p}_i \in \mathbb{R}^2$ is the feature position, $\vec{d}_i \in \mathbb{R}^{128}$ is the corresponding SIFT descriptor, and $z_i$ is the corresponding depth if valid depth data is available from the RGB-D camera.

For the last frame with valid depth information, we record the feature data $\vec{\Theta}^f = \vec{\Theta}$ and the end-effector pose $\pose[0]_c^f = \pose[0]_c$ for later use.

%%%%%%%%%%%%%%%%%%%%%%%%%%%%%%%%%%%%%%% FINAL APPROACH
\subsection{RGB-based final approach phase}\label{sec:rgb-reaching}

\subsubsection{Image-based visual servoing}
An IBVS controller guides a robot to a desired position based on image-plane feature error
\begin{equation} \label{m:error}
\vec{e}(t) = \vec{f} - \vec{f}^*
\end{equation}
where $\vec{f}$ is a set of detected image-plane feature coordinates and $\vec{f}^*$ is a set of \textbf{corresponding} desired image-plane feature coordinates.

The 2-dimensional image-plane velocity of a pixel is related to the 3-dimensional velocity of the camera (rigidly attached to the end-effector) by an image Jacobian  (also known as an interaction matrix)\cite{vs1}
\begin{equation} \label{m:imageJacobian2}
    \dvec{f}_i = \mat{J}_p(\vec{f}, z) \vec{\nu}_c 
\end{equation}
where $\vec{\nu}_c \in{\R^6}$ is the camera spatial velocity, and $\mat{J}_c \in{\R^{2\times 6}}$ is the image Jacobian

\begin{equation} \label{m:imageJacobian3}
\mat{J}_p =
    \begin{pmatrix}
        \dfrac{-{f}}{Z} & 0 & \dfrac{\bar{u}}{Z} & \dfrac{\bar{u}\bar{v}}{-{f}} & \dfrac{-({f}+u^2)}{{f}} & \bar{v} \vspace{0.1cm}\\
        0 & \dfrac{-{f}}{Z} & \dfrac{\bar{v}}{Z} & \dfrac{{f}+\bar{v}^2}{{f}} & \dfrac{-\bar{u}\bar{v}}{{f}} & -\bar{u}
    \end{pmatrix} 
\end{equation}
where $(\bar{u}, \bar{v}) = (u-u_0,v-v_0)$,  $(u_0,v_0)$ is the camera principal point, $(u,v)$ is the pixel location in the image, $f$ is the focal length (in pixel units), and $Z$ is the depth value of the pixel (the distance perpendicular to the image plane). 
For $n \in \mathbb{Z}$ feature points we can stack instances of (\ref{m:imageJacobian2})
\[
\begin{pmatrix} \dot{\vec{f}}_1 \\ \vdots \\ \dot{\vec{f}}_n \end{pmatrix} = 
\begin{pmatrix} \mat{J}_p(\vec{f}_1, z_1) \\ \vdots \\ \mat{J}_p(\vec{f}_n, z_n) \end{pmatrix} \vec{\nu}_c  = \mat{J} \vec{\nu}_c 
\]
where $\mat{J} \in \mathbb{R}^{2n \times 6}$ and $\vec{f}_i, z_i$ are extracted from $\Theta$ in (\ref{eq:features}) and $z_i$ is updated based on forward kinematics 

If $n  \ge 3$ we can solve for the end-effector spatial velocity
\begin{equation}
    \vec{\nu}_e = \mat[e]{J}_c \vec{\nu}_c = -\lambda_i \ \mat{J}^{+} \ \vec{e} 
\end{equation}
where $\lambda_i$ is the gain of the controller, 
$\vec{J}^{+} \in \R^{6\times 2n}$ is the Moore-Penrose pseudo inverse of the stacked image Jacobian from \eq{m:imageJacobian3},
$\mat[e]{J}_c$ is a Jacobian which transforms the spatial velocity of the camera to the end-effector,
$\vec{e} \in \R^{2n}$ is the error vector from (\ref{m:error}).

The depth values used in (\ref{m:imageJacobian3}) can be estimated during the IBVS motion \cite{depth1, depth2} but \cite{vs1} shows that small errors in depth will have a negligible effect on the performance of the controller. In this work, we fix the depth value $Z$ in (\ref{m:imageJacobian2}) at 5\unit{cm}.

\subsubsection{Goal feature prediction}\label{sec:goal-estim}
The IBVS controller requires the image-plane coordinates of the tracked features when the camera is at the grasping pose.
We use information, $\vec{\Theta}^f$ and $\pose[0]_e^f = \pose[0]_e$, from the last frame where depth information is available to estimate this.
We compute the Cartesian coordinates of the SIFT features using \eq{eq:XY}, transform them to the camera pose when the end-effector is at the synthesized grasp pose, then re-project them to the image plane.

\subsubsection{Robust feature matching}
For each subsequent image we compute (\ref{eq:features}), without the depth, and attempt to robustly match the features to $\vec{\Theta}^f$.
%SIFT algorithm is then used on the new image $^{c'}\mat{I}$ to produce a new set of features and descriptors denoted as $^{c'}\vec{f}$.
We use a hierarchy of checks to ensure robust matching:
\begin{enumerate}
    \item distance ratio test outlined in \cite{sift}
    \item duplicate feature removal. SIFT can produce duplicate features with different scale and orientation, so we remove any features within 5 pixels of another match (where the higher quality match remains).
    \item loop constraint\cite{lcc}
    \item the fundamental matrix is calculated using RANSAC. This produces a list of inlier, and outlier matches where only inlier matches are retained \cite{ransac}.
    \item A 20 $\times$ 20 grid is placed over the image, where a maximum of one matched feature is kept per grid cell. This ensures that the Jacobian in \eq{m:imageJacobian2} is well conditioned and that feature points are well spread across the image.
\end{enumerate}

\begin{figure*}[t]
    \centering
    \begin{subfigure}{0.3\textwidth}
        \centering
        \includegraphics[width=5.3cm, height=3cm]{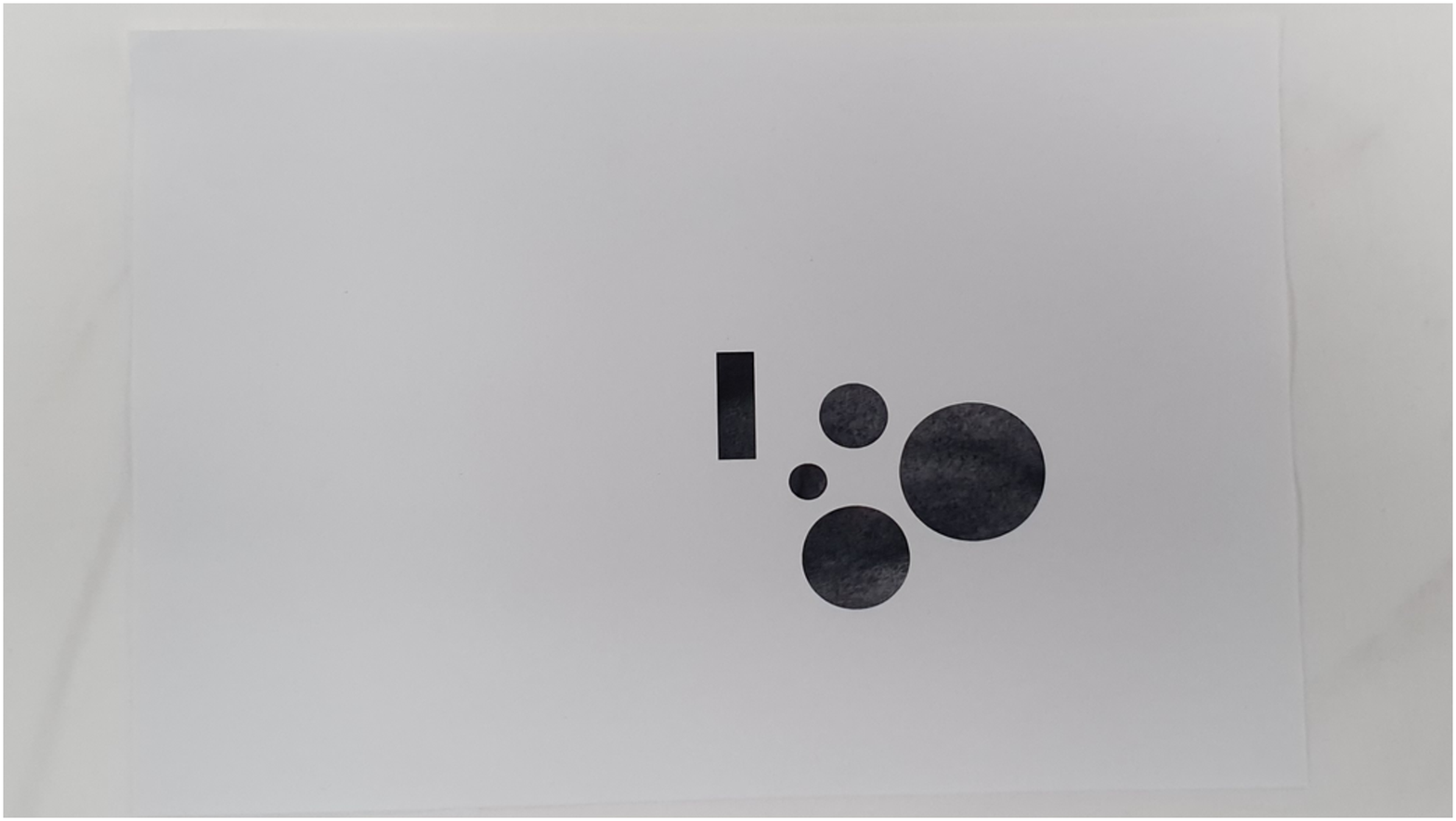}
        \caption{}
        \label{fig:target}
    \end{subfigure}
    \hfill
    \begin{subfigure}{0.3\textwidth}
        \centering
        \includegraphics[width=5.3cm, height=3cm]{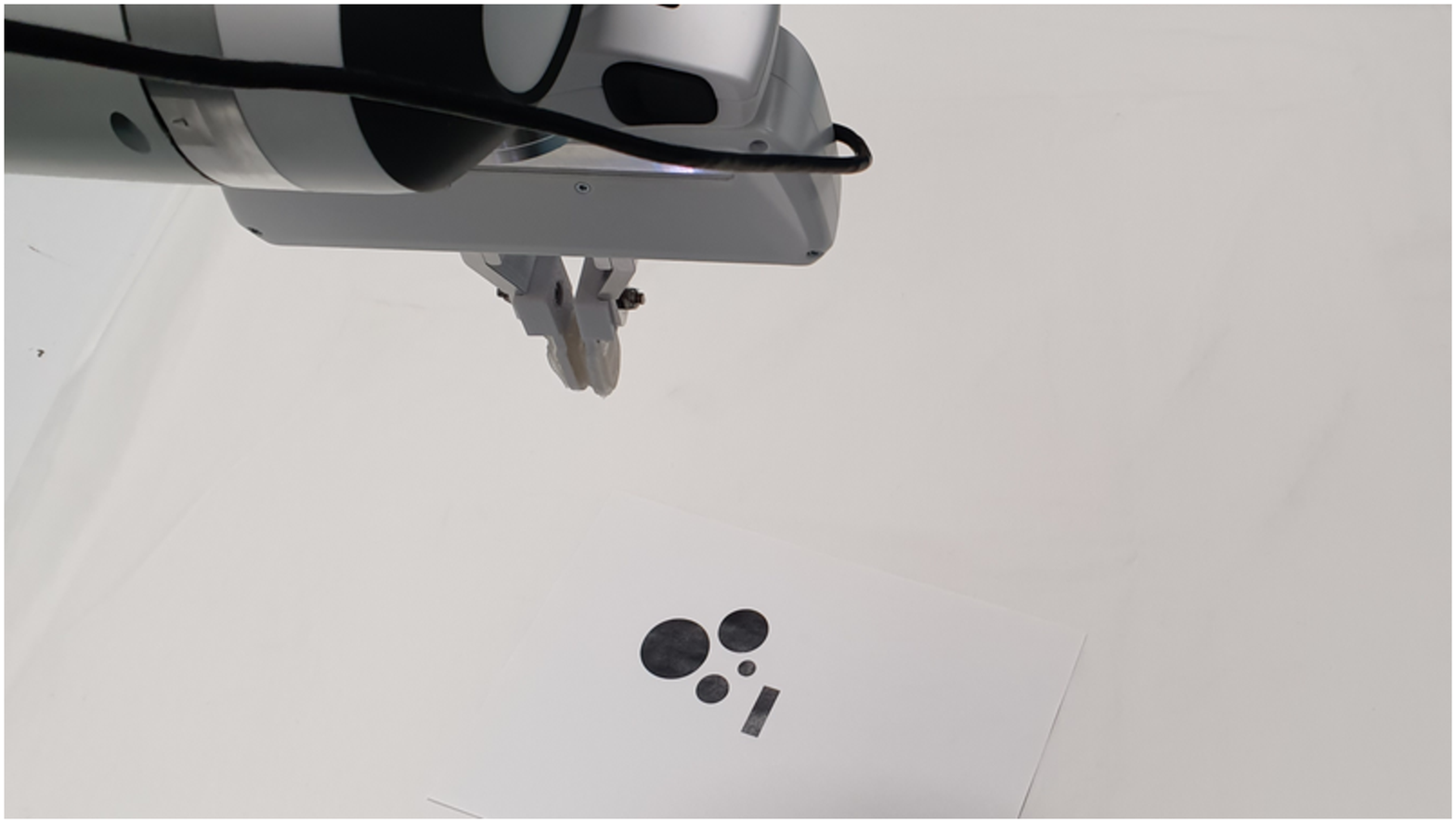}
        \caption{}
        \label{fig:initial}
    \end{subfigure}
    \hfill
    \begin{subfigure}{0.3\textwidth}
        \centering
        \includegraphics[width=5.3cm, height=3cm]{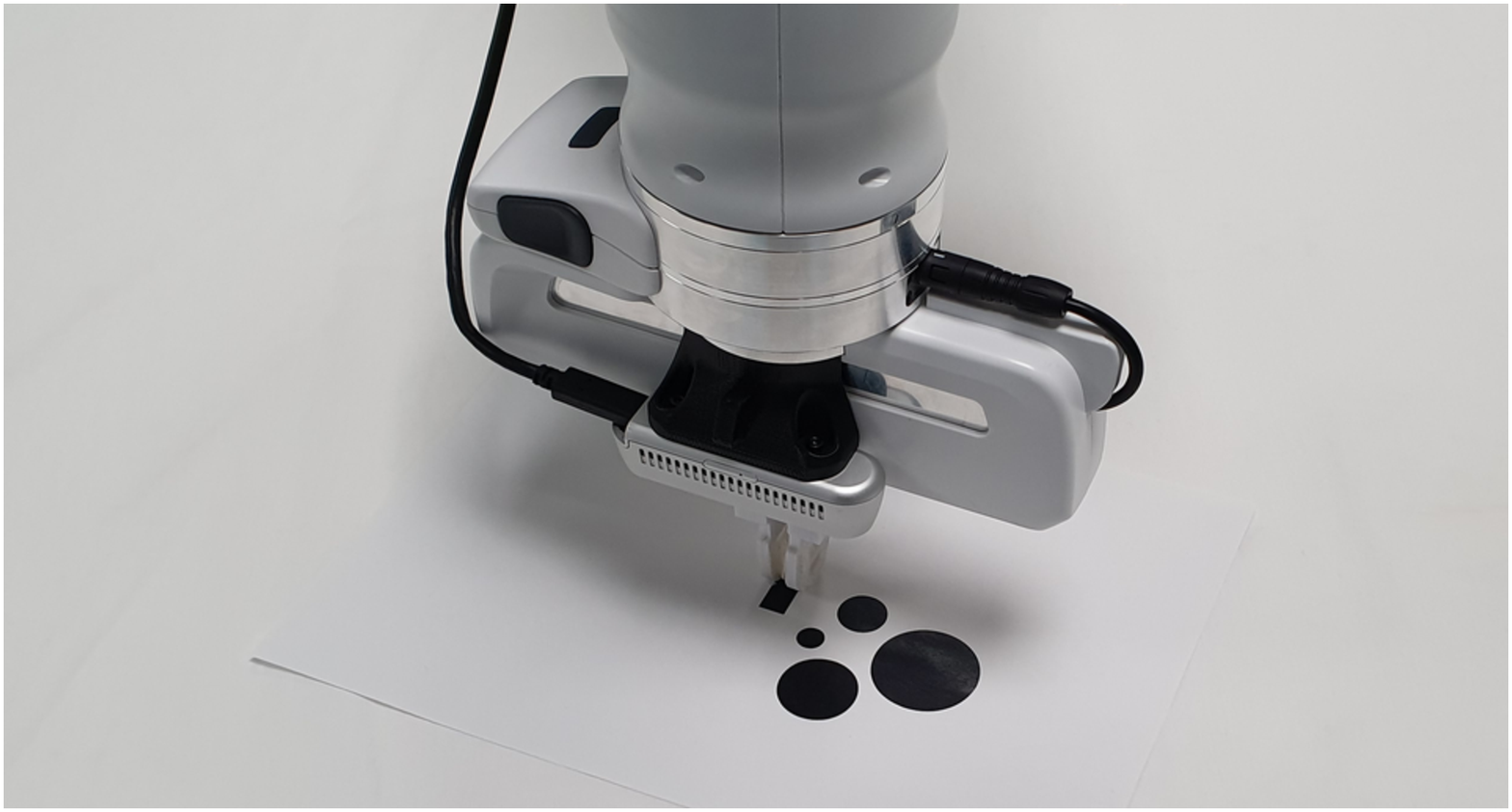}
        \caption{}
        \label{fig:at goal}
    \end{subfigure}
       \caption{(a) The target features printed on A4 paper. (b) The robot's position before attempting to reach the goal. (c) the robot in the goal state.}
       \label{fig:three graphs}
\end{figure*}

\begin{figure*}[t]
    \centering
    \begin{subfigure}{0.55\textwidth}
        \centering
        \includegraphics[width=8.5cm,height=3.22cm]{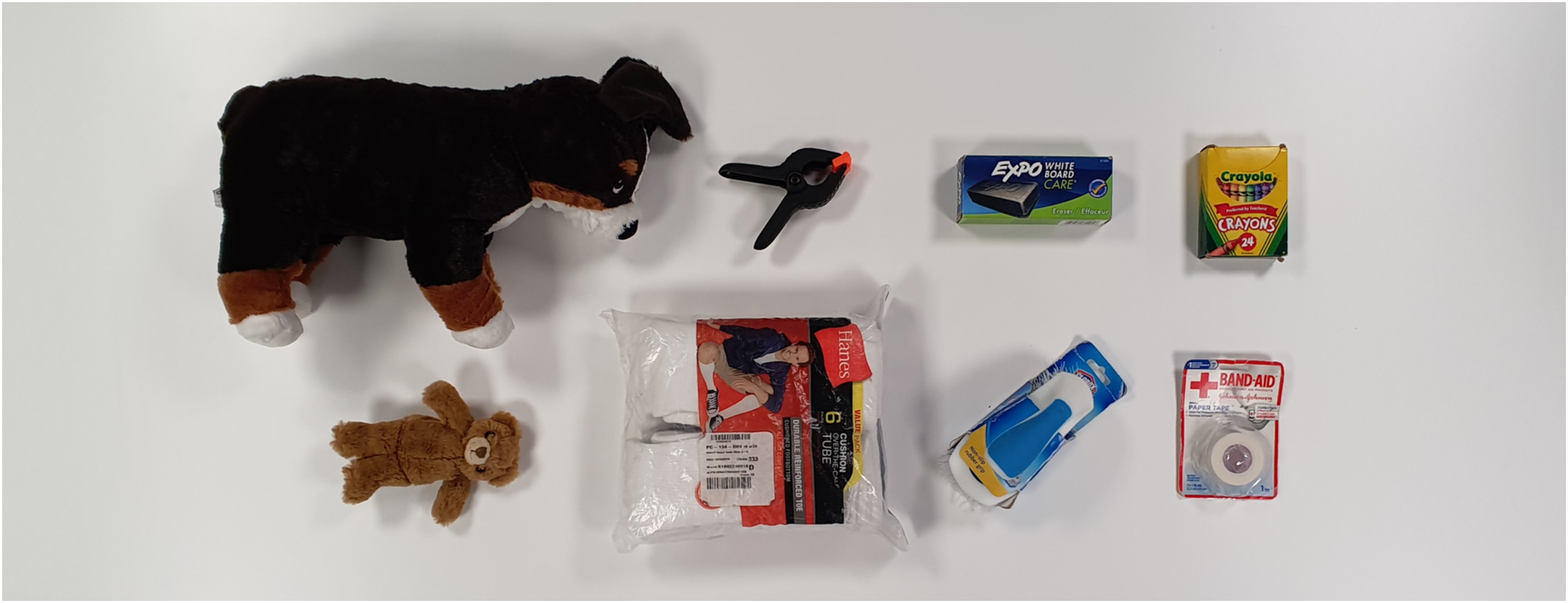}
        \caption{}
        \label{fig:objects}
    \end{subfigure}
    \hfill
    \begin{subfigure}{0.44\textwidth}
        \centering
        \includegraphics[width=6cm, height=3.22cm]{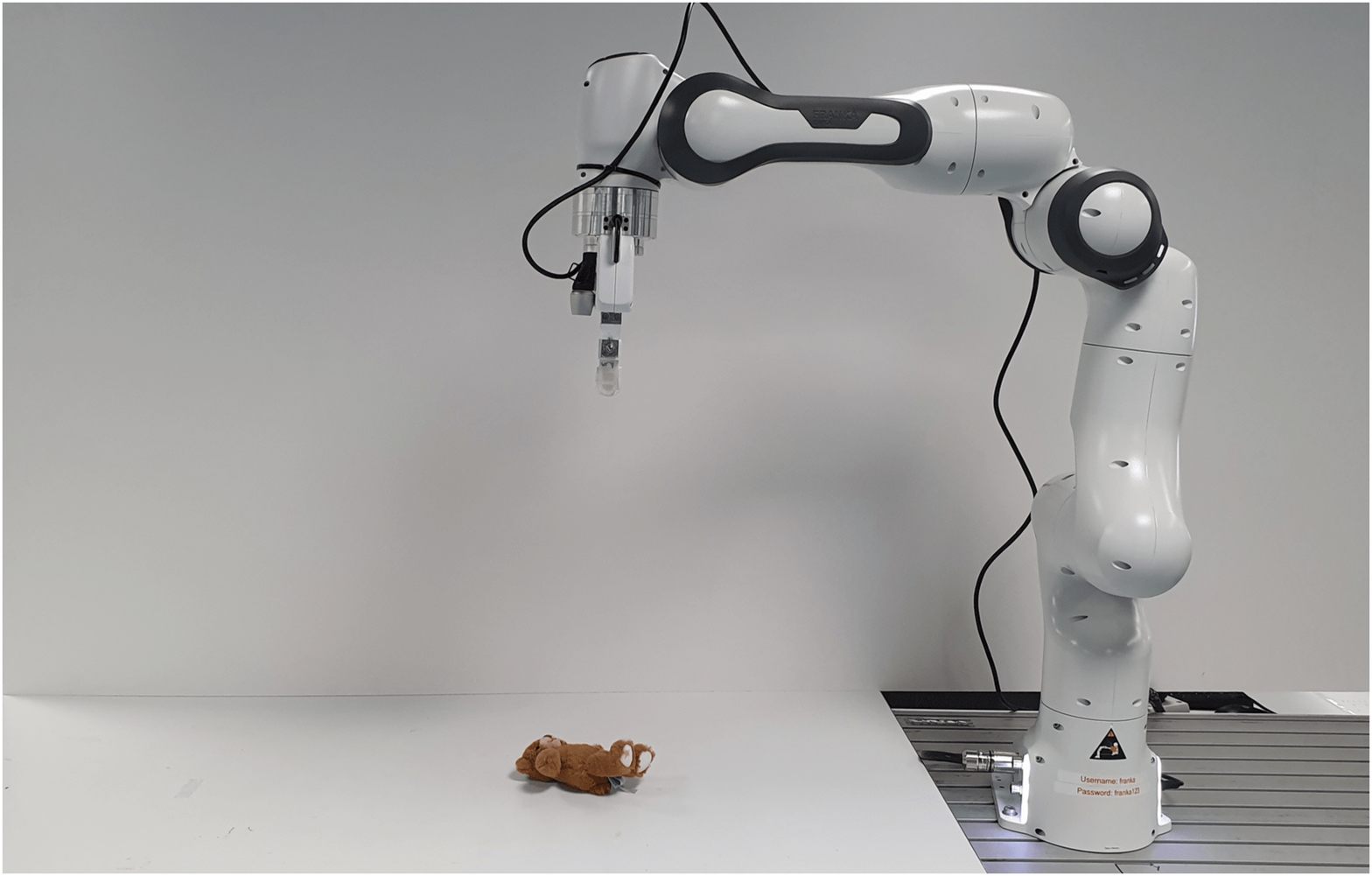}
        \caption{}
        \label{fig:grasp initial}
    \end{subfigure}
       \caption{(a) The 8 household objects which are to be grasped. (b) The robot's position before attempting a grasp.}
       \label{fig:three graphs 2}
\end{figure*}

\subsection{Switching VS Control Scheme}\label{sec:switching}
The RealSense D15 camera has a rated minimum sensing distance of 16\unit{cm} which agrees with our experience.
Violation of minimum distance results in depth values of NaN.
Rather than counts NaNs in the image and choose a threshold we adopt a simple and conservative strategy that deems range data invalid when the object is sensed to be within 25\unit{cm} of the camera. 

Our controller uses PBVS when range data is available and IBVS when it is not.
This allows us to exploit the benefits of each, while avoiding their major shortcomings. PBVS provides an optimal Cartesian path to the goal but is prone to errors introduced from camera calibration and robot odometry. This makes PBVS best suited to getting the robot close to the goal. IBVS is very robust to sensor error but may produce sub-optimal Cartesian paths and is best utilized in the final approach. Figure \ref{fig:switching}  demonstrates this approach.
A simple filter ensure continuity of velocity at the transition.

\subsection{Grasping}

When the error calculated in (\ref{m:error}) becomes sufficiently small, the servoing is considered complete and a grasp can be attempted. This is completed by instructing the fingers of the robot's grippers to close.

\begin{figure*}[t]
    \centering
    \includegraphics[width=1\textwidth]{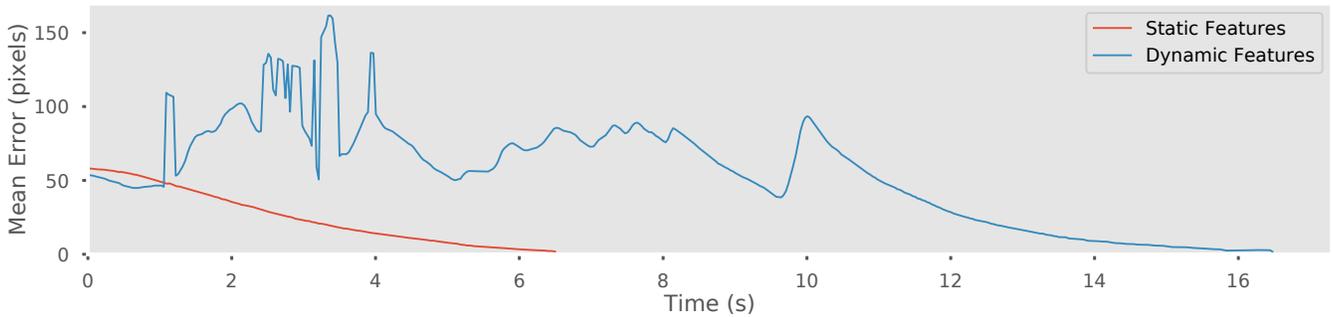}   
    \caption{Experiment 1: Target Feature Error in Static and Dynamic Test with Known Features}
    \label{fig:static test}
\end{figure*}

\begin{figure*}[t]
    \centering
    \includegraphics[width=1\textwidth]{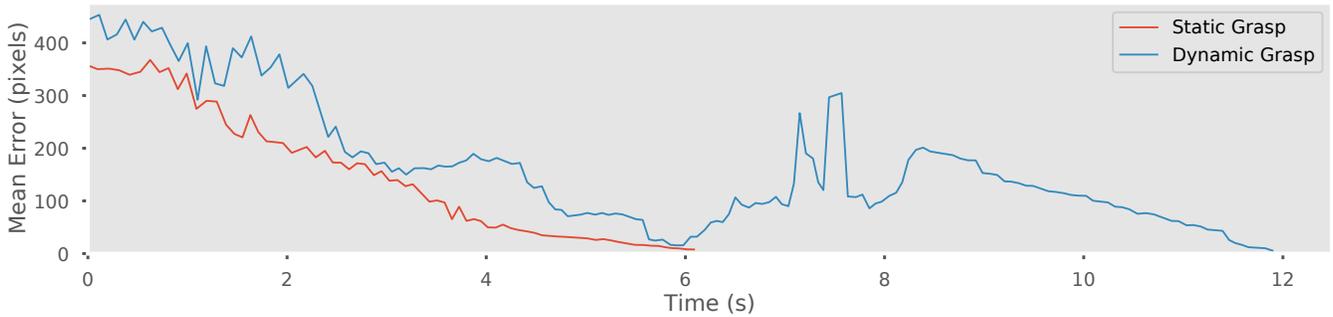}   
    \caption{Experiment 2: Target Feature Error in a Static and Dynamic Grasping Trial}
    \label{fig:ob test}
\end{figure*}

\begin{figure}[t]
    \centering
    \begin{subfigure}{0.49\columnwidth}
        \centering
        \includegraphics[width=1\textwidth]{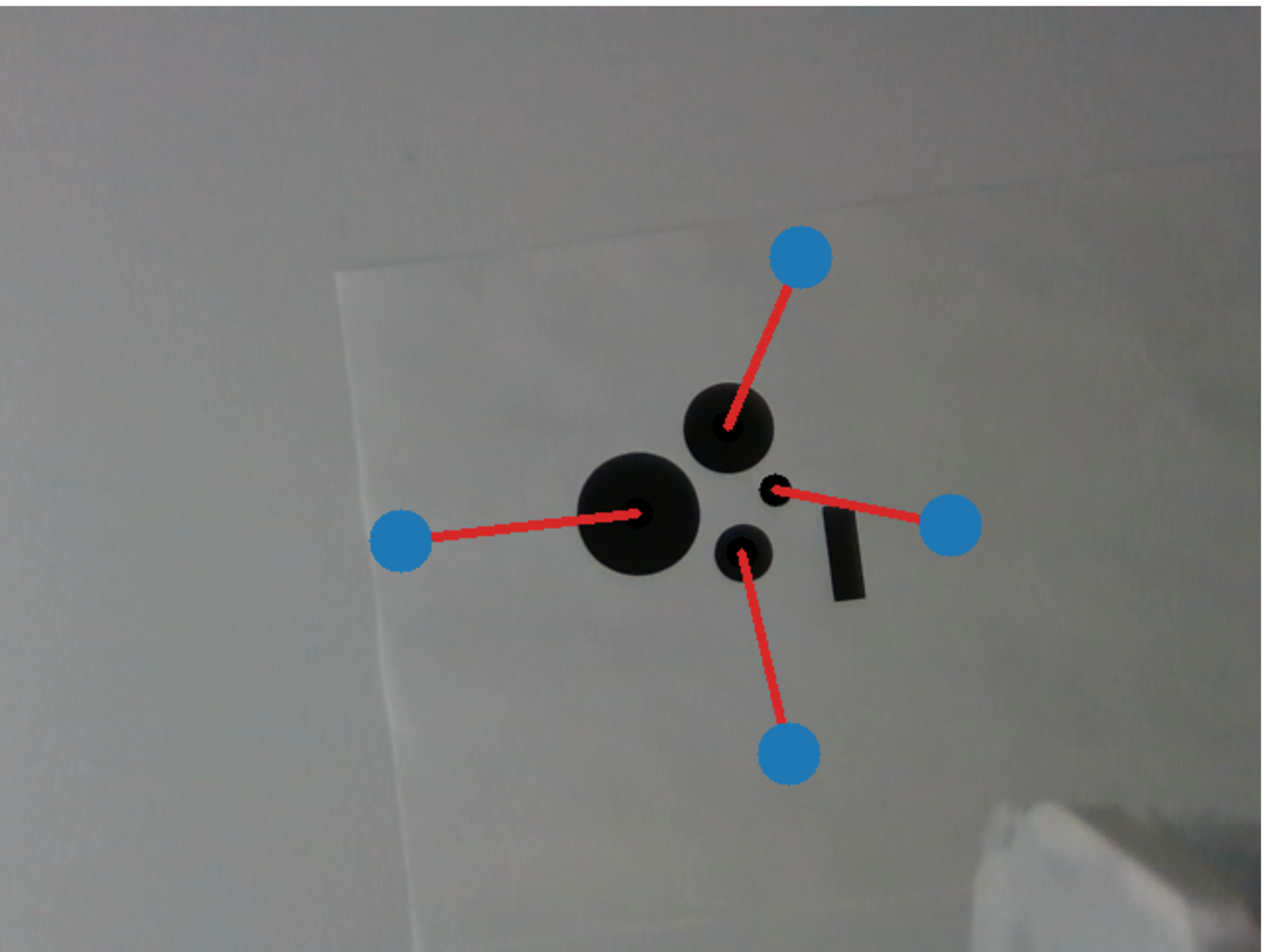}
        \caption{}
        \label{fig:static test 3}
    \end{subfigure}
    \hfill
    \begin{subfigure}{0.49\columnwidth}
        \centering
        \includegraphics[width=1\textwidth]{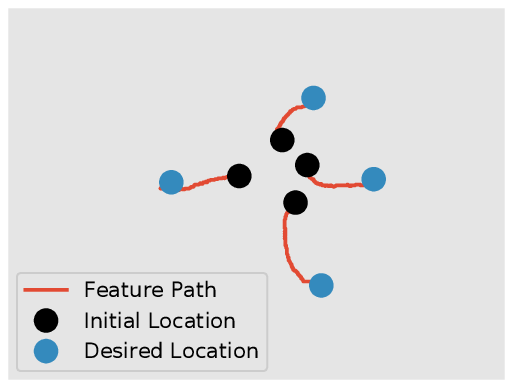}
        \caption{}
        \label{fig:static test 2}
    \end{subfigure}
       \caption{Experiment 1: Test with Binary Features (a) Photo of Initial and Desired Feature Configuration from the Initial Camera POV. (b) Actual Target Feature Trajectory in the Image Plane.}
       \vspace{-14pt}
       \label{fig:static tests}
\end{figure}

\section{Experiments}\label{sec:exp}

We validate and evaluate  our approach through testing on an arm-type robot. Our approach is realized using ROS middleware and primarily Python code. Our experiments first seek to validate that the approach works given  ideal conditions, before evaluating the robustness and consistency on repeated grasping trials.

\subsection{Equipment}

As shown in Figure \ref{fig:cover}, all experiments are performed using a Franka-Emika Panda robot, equipped with 3D-printed grippers using a design from \cite{gripper}. We use an Intel RealSense D415 camera to provide RGB and depth information, which is mounted to the robot's end-effector.

\subsection{Experiment 1: Predicting Target Feature Configuration}
We first validate our target feature configuration predictor through an IBVS controller with both static and dynamic targets. These tests seek to demonstrate that the target feature configuration can be predicted using only initial depth information, to enable closed-loop control.

To remove any potential unreliability due to feature matching we print a simple planar target with blobs of different sizes and shapes which we analyze with classical binary vision techniques, see  Figure \ref{fig:target}.  The goal of the robot is to place its end-effector at the reference pose of this target, see Figure \ref{fig:at goal}.

In the static test, the robot is initialized 40~cm above the table, see Figure \ref{fig:initial}, with the target having a random pose and located within the camera's field of view. In the dynamic tests, the target is moved by hand in a random translational motion such that they remain visible to the camera.

\subsection{Experiment 2: Grasping Trials}

In this experiment we evaluate our switching visual servoing scheme on grasping tasks with common household objects and use SIFT features. We perform dynamic grasping trials where the object on the table is moved in a random fashion while the robot attempts to grasp it. Some of the target objects are displayed in Figure \ref{fig:objects} and vary in size, and grasp difficulty. In each test, one of the objects is  placed randomly on a table located within a 30 $\times$ 30cm zone.
The starting configuration of the robot with an object is shown in Figure \ref{fig:grasp initial}.
The robot then attempts to grasp the object ten times while the object is being moved.

\addtolength{\textheight}{-0.12cm}   % This command serves to balance the column lengths

\section{Results}\label{sec:results}

The results from the Experiment 1 static test show that, given perfect correspondence between features, the approach will allow the robot to achieve the goal. Figure \ref{fig:static test} shows the average feature error between each feature point and the desired location of that feature point in the image plane. 
Figure \ref{fig:static test 3} displays the initial view of the features from the camera's point of view, the desired feature configuration as predicted by our algorithm and the ideal IBVS path to be followed. 
Figure \ref{fig:static test 2} displays the feature path actually taken on the image plane. The paths are close to straight but deviation is due to the fixed depth value used to compute the image Jacobian.

The results from the Experiment 1 dynamic test verifies that the controller can operate with dynamic scenes. Figure \ref{fig:static test} displays the target feature error. We observe large upward spikes when the target is moved in the scene but the controller continues to drive the features to their goal configuration.

Results for Experiment 2 dynamic grasping are shown in Figure \ref{fig:ob test} which displays the feature-point error in a static and dynamic object grasp attempt. We observe that there is more noise present in the system when using SIFT features compared to the simple binary features of Figure \ref{fig:static test}, however the robot consistently reaches its goal.

\newcolumntype{M}[1]{>{\centering\arraybackslash}m{#1}}
\begin{table}[t]
    \centering
    \renewcommand{\arraystretch}{1.3}

    \caption{Experimental Results From Grasping Trials}
    \label{tab:results}

    \begin{tabular}{ c | M{2cm} }
    % \begin{tabular}{l | r} }
    \hline
    & \multicolumn{1}{c}{\textbf{Grasp Rate}} \\
    \hline\hline
    Static Objects \cite{doug} & 92\%\\
    Dynamic Objects \cite{doug} & 0\%\\
    \hline\hline
    Our Approach with Dynamic Objects & \textbf{76.25\%}\\
    \hline
    \end{tabular}

\end{table}

Summary results from the repeated dynamic grasping trials are displayed in Table \ref{tab:results}. The results in  rows 1 and 2 are taken directly from \cite{doug} and highlight the issue that a closed-loop grasper has for the case of moving objects due to the camera minimum sensing distance.

In contrast, our switching controller (row 3) shows a significant increase in performance for closed-loop grasping of moving objects with a grasping success rate is 76.25\%. 
 The performance of our system was stronger on larger objects and objects which had many unique SIFT features. This is expected since our approach relies on numerous unique features being present in the scene with enough remaining to be visible to the camera at the grasp point. While this limits the types of scenes and objects our approach will reliably work on, it could be mitigated through alternative or additional features such as lines, object shapes, or image moments.

The main failure mode was with objects that moved fast and left the camera's field of view. This could be mitigated by incorporating an object velocity estimator and feed-forward control.
Other failure modes included blurry images due to extreme object velocity, and some weakness in the use of SIFT features such as stability of feature position and descriptor over very large changes of scale.

\section{CONCLUSIONS}
Robotic grasping controllers that are not closed-loop will fail to grasp moving objects.
Closed-loop grasp controllers based on RGB-D imagery, such as \cite{doug}, can track a moving object, but fail when the sensor's minimum object distance is violated just before grasping.

This paper has shown how image-based visual servoing can improve the performance of closed-loop RGB-D-based grasping algorithms for the case of moving objects. 
We achieve this by servoing toward image-plane goal features that are \textit{predicted} from a  depth image, robot encoder-based pose, and the grasp synthesiser's goal pose.
This is quite different to most previous IBVS work where the goal feature configuration is assumed to be known. 
Using IBVS in this way retains all the advantages of the RGB-D-based grasp synthesizer such as not requiring a model of the object being grasped.

We have demonstrated the robustness of this new approach in the context of dynamic closed-loop grasping and shown a greatly improved grasp success rate.

%\section*{ACKNOWLEDGMENT}

% \balancecolsandclearpage

\bibliographystyle{IEEEtran} 

% \bibliography{IEEEabrv, ref} % For normal
\bibliography{ref} % For overleaf

\end{document}